\RequirePackage{fix-cm}
\documentclass[onecolumn,10pt]{svjour3}
\smartqed  % flush right qed marks, e.g. at end of proof
\usepackage{graphicx}
\usepackage{fixltx2e}
\usepackage[misc]{ifsym}
\usepackage{float}
\usepackage[multi-part-units=single]{siunitx}
\usepackage[font=small,labelfont=bf,tableposition=top]{caption}

\DeclareCaptionLabelFormat{andtable}{#1~#2  \&  \tablename~\thetable}
\begin{document}

\title{Towards Automatic Lesion Classification in the Upper Aerodigestive Tract Using OCT and Deep Transfer Learning Methods}


\titlerunning{Lesion Classification in the Upper Aerodigestive Tract}        % if too long for running head

\author{Nils Gessert$^1$        \and
        Matthias Schl\"uter$^1$  \and
        Sarah Latus$^1$  \and
        Veronika Volgger$^2$ \and
        Christian Betz $^3$ \and
        Alexander Schlaefer$^1$  %etc.
}


\institute{\Letter \quad Nils Gessert, \email{nils.gessert@tuhh.de}, Tel.: +49 (0)40 42878 3389, https://orcid.org/0000-0001-6325-5092 \\ \\ $^1$ Hamburg University of Technology, Hamburg, Germany \\ $^2$ Ludwig-Maximilians-Universität München, Clinic and Polyclinic for Otolaryngology, Munich, Germany \\ $^3$ University Medical Center Hamburg-Eppendorf, Clinic and Polyclinic for Otolaryngology, Hamburg, Germany}

\date{Preprint, submitted to CARS 2019, accepted for publication}

\maketitle

\begin{abstract}

Early detection of cancer is crucial for treatment and overall patient survival. In the upper aerodigestive tract (UADT) the gold standard for identification of malignant tissue is an invasive biopsy. Recently, non-invasive imaging techniques such as confocal laser microscopy and optical coherence tomography (OCT) have been used for tissue assessment. In particular, in a recent study experts classified lesions in the UADT with respect to their invasiveness using OCT images only. As the results were promising, automatic classification of lesions might be feasible which could assist experts in their decision making. Therefore, we address the problem of automatic lesion classification from OCT images. This task is very challenging as the available dataset is extremely small and the data quality is limited. However, as similar issues are typical in many clinical scenarios we study to what extent deep learning approaches can still be trained and used for decision support.

\end{abstract}

\section{Methods} 

\begin{figure*}
  \centering
  \includegraphics[width=1\textwidth]{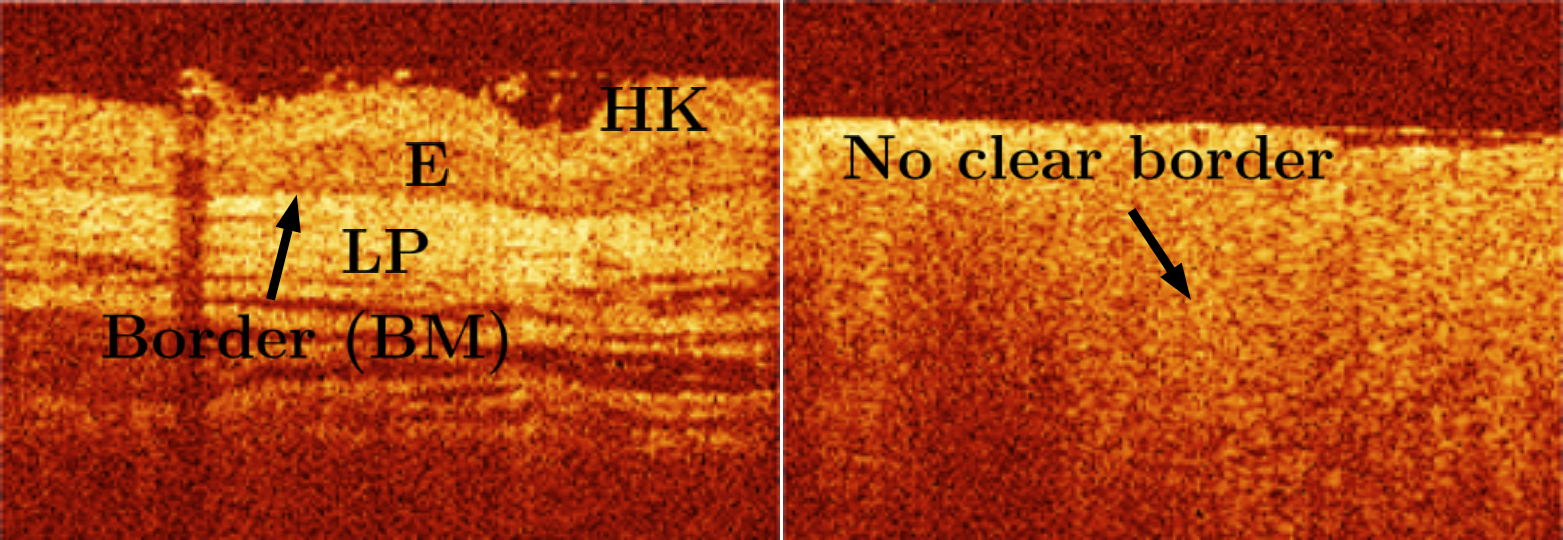}
\caption{Left, a hyperkeratotic lesion at the floor of the mouth is shown. BM denotes the basement membrane, LP denotes the lamina propria, E denotes the epithelial layer and HK denotes hyperkeratosis. Right, an invasive lesion at the floor of the mouth is shown. There is no clear border between the epithelial layer and the lamina propria.}
\label{fig:example}       % Give a unique label
\end{figure*}

We use the data set from the study performed by Volgger et al. \cite{Volgger2013} which contains 100 lesions with 91 benign and 9 invasive lesions. Example images are shown in Figure~\ref{fig:example}. An invasive lesion is characterized by the absence of a clear border between the epithelial layer and the lamina propria. This notable difference should be a learnable feature from the images. The binary ground-truth labels were obtained by histological evaluation of a biopsy taken after OCT imaging. Due to the small size of the data set, we use a form of leave-one-invasive-lesion-out cross-validation, i.e., we construct 8 validation splits with 1 invasive and 9 benign lesions each. We employ the state-of-the-art deep learning models Resnet18, Densenet121 and SE-ResNeXt50 which are pretrained on ImageNet for transfer learning which has been successful for OCT classification tasks \cite{gessert2019automatic}. The OCT images are resized from 260x180 pixels to 224x224 pixels using bilinear interpolation to match the models’ standard input size. The classification layer is replaced by a layer with two outputs. We consider transfer learning with fine-tuning of all weights and a variant where we only retrain the classifier. For comparison, we also train the models from scratch. We employ online data augmentation with random flipping along the horizontal dimension and we apply random changes in brightness, contrast and saturation. Furthermore, we use regularization with dropout ($p=0.2$) and a small batch size ($b=5$), which induces more variation during optimization. We use Adam for optimization with a very small learning rate of $\num{10e-6}$. As the two classes are highly unbalanced, we employ loss balancing with the normalized inverse class frequency as a weight factor in the loss function. For evaluation, we report the sensitivity, specificity, accuracy and F1-score.

\section{Results}

\begin{table}
\centering
\caption{Results in percent for different training scenarios and models compared to the average human rater in \cite{Volgger2013}. We compare training from scratch (SCR), transfer learning with fine-tuning (FT) and retraining of only the classifier (RC). The values are the mean for leave-one-invasive-lesion-out cross-validation.}
\label{tab:result}
	\begin{tabular}{l l l l l}
	 & Accuracy & Sensitivity & Specificity & F1-Score \\ \hline \\
	 Densenet121 SCR & $60.18$ & $37.50$ & $69.63$ & $61.00$ \\
	 Densenet121 FT & $64.10$ & $50.00$ & $58.17$ & $62.74$ \\
	 Densenet121 RC & $77.10$ & $75.00$ & $77.32$ & $73.18$ \\ \hline \\
	 Resnet18 SCR & $60.90$ & $50.00$ & $60.98$ & $60.06$\\
	 Resnet18 FT & $55.63$ & $62.50$ & $54.25$ & $52.01$ \\
	 Resnet18 RC & $76.10$ & $75.00$ & $79.65$ & $75.42$ \\ \hline \\
	 SE-Resnext50 SCR & $60.19$ & $37.50$ & $68.17$ & $62.09$ \\
	 SE-Resnext50 FT & $66.19$ & $62.50$ & $69.73$ & $64.27$ \\
	 SE-Resnext50 RC & $82.76$ & $75.00$ & $84.95$ & $81.98$ \\ \hline \\
	 Human Rater & - & $81.50$ & $72.50$ & - \\ \hline \\	 	 	 
	\end{tabular}
\end{table}

The results are shown in Table~\ref{tab:result}. We compare to the mean sensitivity and specificity of human raters reported in \cite{Volgger2013}. As expected, training from scratch is not very successful as the performance across all metrics is low with an accuracy of $\SI{60}{\percent}$ and a sensitivity of $\SI{37.50}{\percent}$-$\SI{50}{\percent}$. Also, transfer learning with fine-tuning of all weights only improves performance marginally. Retraining the classifier only achieves a reasonable performance with a sensitivity and specificity of $\SI{75.00}{\percent}$ and $\SI{82.76}{\percent}$ for SE-ResNeXt50 which is similar to the average human rater with a sensitivity of $\SI{81.50}{\percent}$ and a specificity of $\SI{72.50}{\percent}$. All three architecture variants perform similar across the different training scenarios and the highest sensitivity is achieved by the average human rater. It is notable that SE-ResNeXt50 performs best for this problem while also being the most recent and best performing model on ImageNet out of the three models we chose.

\section{Conclusions}

We consider the problem of automatic lesion classification in the UADT using OCT images and deep learning despite facing the challenge of a very small dataset. We tackle this issue with transfer learning techniques, data augmentation and regularization. Our results show that a model where only the classifier is retrained performs similar to the average human rater from a previous study. Training the model from scratch and also transfer learning with fine-tuning of all weights, a typical approach for medical image classification \cite{gessert2019automatic}, fails in this case. This is likely caused by the small dataset size. Our results with a retrained classifier indicate that OCT combined with deep learning methods might be useful to assist experts in their decision making when assessing lesions in the UADT. This holds true for three recent architectures. Moreover, a larger dataset of high quality full resolution OCT images will likely improve the classification performance.

%\begin{acknowledgements}
%\end{acknowledgements}

\bibliographystyle{spmpsci} 
\bibliography{egbib}   % name your BibTeX data base

\end{document}